\documentclass{article}
\usepackage{spconf,amsmath,graphicx}
\usepackage{gensymb}
\usepackage{colortbl}
\usepackage{acronym}
\newacro{CNN}[CNN]{convolutional neural network}
\newacro{CCM}[CCM]{Color Conversion Matrix}
\newacro{RAE}[RAE]{recovery angular error}
\title{Probabilistic Color constancy}
\name{Firas Laakom$^{\dagger}$, Jenni Raitoharju$^{\dagger \mathsection}$,  Alexandros Iosifidis$^{\star}$, Uygar Tuna$^{\ddagger}$, Jarno Nikkanen$^{\ddagger}$, Moncef Gabbouj$^{\dagger}$  }
\address{$^{\dagger}$Tampere University, Faculty of Information Technology and Communication Sciences, Finland  \\  
$^{\mathsection}$ Finnish Environment Institute, Programme for Environmental Information, Finland \\
    $^{\star}$Aarhus University, Department of Engineering,  Denmark \\
     $^{\ddagger}$ Xiaomi, Finland
    }

\begin{document}
%\ninept
%
\maketitle
\begin{abstract}
In this paper, we propose a novel unsupervised color constancy method, called Probabilistic Color Constancy (PCC). We define a framework for estimating the illumination of a scene by weighting the contribution of different image regions using a graph-based representation of the image. To estimate the weight of each (super-)pixel, we rely on two assumptions: (Super-)pixels with similar colors contribute similarly and darker (super-)pixels contribute less. The resulting system has one global optimum solution. The proposed method achieves competitive performance, compared  to the state-of-the-art, on INTEL-TAU dataset.

% The results are presented and discussed. The source code website is provided in Section Experimental Results.
\end{abstract}
\section{Introduction}
\label{sec:intro}
In the image processing pipeline of every camera, the impact of illumination on the image colors is attenuated. This well-established computer vision problem is usually referred to as color constancy \cite{maloney1986color}. Discounting the effect of the illumination and restoring the `true' colors of objects in the scene improves the subjective image quality and it is an essential building block in several higher-order image processing systems, such as object tracking and object classification \cite{gijsenij2011computational}.  It allows objects and scene content to have consistent color distribution under different lighting conditions. Thus, they are easier tracked,  classified, and recognized. Endowing digital cameras with this ability is a challenging and ill-posed problem. Computational color constancy approaches usually rely on a two-step process. In the first step, the global illumination is estimated and, in the second step, all color pixels of the scene are normalized using the estimated illuminant color. Thus, the color constancy problem is equivalent to illumination estimation. 
 
Recently, with the advancement of machine learning techniques in general and deep learning in particular, several neural network-based illumination estimation techniques have been proposed \cite{22,44,mine,mine2}. These methods have usually led to state-of-the-art results over several datasets. However, they require calibrated images with known ground-truth illumination to fine-tune their parameters and creating such images for a given sensor is time-consuming. Besides, supervised approaches rely on the assumption that the inference of a given scene colors distribution and illumination conditions are the same in the training and testing phases. Thus, they tend to over-fit to training data and are sensitive to camera models and scenes  encountered during the training phase \cite{mine}.  We thus argue there is still interest in unsupervised methods as they usually do not require any training data and are independent of the acquisition device. Furthermore,  unsupervised methods are computationally fast and have small memory and power footprint compared to deep learning approaches.  Thus, they are more convenient especially for low-power devices, such as mobile phones. 

Unsupervised methods usually make assumptions about the nature of the colors and statistical properties of the scene to estimate illumination \cite{3}. The Gray world algorithm \cite{4}, for example, assumes that illumination color is the average of colors present in a scene, while the White-Patch \cite{2} uses the maximum RGB responses in a scene as an estimate for illumination. In \cite{yang2015efficient} and \cite{qian2019finding}, the problem of color constancy is reformulated as the search of grey pixels, i.e., pixels representing an achromatic surface and using the cast of the grey pixels as an illumination  estimate. Most of these methods assume that the illumination color is one of the scene pixel colors. In this paper, we relax this constraint and assume that the color of the illumination belongs to the convex hull of the pixel colors present in the scene. Thus, it can be obtained as a convex combination of the scene colors.  We propose a probability-based formulation defined on the color similarity graph of an image to obtain such a solution.

Graphs offer a rich and compact representation of the spatial and color-wise similarity relations between pixels/regions in an image. They allow to model constraints and properties of a particular problem in a single framework solving a joint optimisation problem. They have been used to model many tasks in computer vision, such as saliency estimation \cite{aytekin2018probabilistic}, object tracking \cite{malcolm2007multi}, and image segmentation \cite{vicente2008graph,aytekin2017learning,shi2000normalized}.  In this paper, we model the color constancy problem under a probabilistic framework encoding the illumination assumptions and smoothness constraints into an optimization problem. We rely on three key assumptions to model the connectivity between (super-)pixels \cite{achanta2010slic} and the constraints used to define the optimization process. The first assumption is that (super-)pixels with similar colors encode the same information about the global illumination. The second assumption is that darker pixels encode less information about the scene illumination than the brighter ones. The third assumption is that the scene illumination can be obtained by a weighted sum of the RGB colors of the pixels in the scenes. The first two constraints enable us to define the connectivities of the graph. The adopted optimization scheme has a closed-form global optimum solution, which estimates the color of the light source in a scene.

The main contributions of the paper are as follows:
\begin{itemize}
\item We propose a novel probability-based unsupervised color constancy method, called Probabilistic Color Constancy (PCC). Following this approach, we formulate a novel optimization problem based on the color similarity graph of an image.

\item The proposed approach was tested on the INTEL-TAU dataset, where it achieved comparable results with state-of-the art unsupervised color constancy  methods.

\item The proposed unsupervised approach can be used as a generic framework for color constancy approaches, where different prior information can be integrated in the future.

\end{itemize}

% The paper is organized as follows. Section 2 gives a global overview-of related works. Sections 3 provides a description of the proposed framework. Section 4 provides experimental results on two different color constancy datasets. Finally, Section 5 draws the conclusions.

\section{Related work}
\label{sec:related}
 
\subsection{Color Constancy} 
The aim of color constancy methods is to estimate the global illumination of a scene in order to correct the colors of an image representing the scene.  Several unsupervised methods were grouped into a single framework \cite{5}, where the illumination estimate $\textbf{I}^{est}$ is expressed as follows: 
\begin{equation}
                           \textbf{I}^{est}(n,p, \sigma) =  \frac{1}{k}  \left(\int_x \int_y | \bigtriangledown^n  \rho_{\sigma}(x,y) |^p dxdy\right)^{ \frac{1}{p}},
\end{equation}
where $n$  denotes the derivative order, $p$ the order of the Minkowski norm, and $k$  the normalization constant for $\textbf{I}^{est}$. Also, $\rho_{\sigma}(x,y) = \rho(x,y) * g_{\sigma}(x,y) $ denotes the image convolution, .ie., $\rho(x,y)$, with a Gaussian filter $g_{\sigma}(x,y)$ having a scale parameter $\sigma$. This framework allows for deriving different algorithms simply by setting appropriate values for $n$, $p$ and $\sigma$. The Gray-World method \cite{4} corresponds to $(n = 0,p =1, \sigma = 0)$. It corresponds to the average of all the pixel values in the scene. The White-Patch \cite{2}  corresponds to $(n = 0,p = \infty, \sigma = 0)$. It considers the pixel with the maximum response on each channel as the main estimate of the illumination. The Gray-Edge methods \cite{5} can be derived by using $(n = 1,p = p, \sigma =\sigma  )$. The input image can by subsequently corrected by
dividing each pixel channel-wise with the estimated illumination $\textbf{I}^{est}$.

\subsection{Graph-based learning} 
% Graph-based approaches have been heavily used in computer vision tasks such as classification \cite{iosifidis2016multi,bai2012graph}, semi-supervised learning \cite{camps2007semi}, tracking \cite{gomila2003graph} and segmentation \cite{vicente2008graph,aytekin2017learning}. The works closest to the proposed method fall inside image segmentation \cite{vicente2008graph,felzenszwalb2004efficient,aytekin2017learning}. The main task is to divide an image into different regions possessing similar properties.  Graph-cut \cite{vicente2008graph} Constructs a pathway connecting all the edges to cut the graph.  In order to cut the graph from a start node to an end node, it minimize the total weight sum of the path. As a result, the edges E are set to reflect the cost of each local path. However, the minimum cut criteria often lead to cutting isolated nodes in the graph due to the small values achieved by partitioning such nodes. To tackle this problem, Normalized-cut \cite{shi2000normalized}  was proposed which focuses on extracting the global impression of an image rather than focusing on local features and their consistencies in the image data. Normalized-cut criterion takes into consideration both the total dissimilarity between the different splits as well as the total similarity within the splits. It considers the fraction of the total edge connections to all the nodes in the graph as the cost function. In \cite{aytekin2018probabilistic}, a probabilistic graph based framework is proposed for salient object detection.

Graph-based approaches have been successfully used to solve several computer vision tasks such as image segmentation based on semantic information \cite{vicente2008graph,aytekin2017learning,shi2000normalized} and weighting pixels of an image based on their saliency estimated from the color and proximity in the image \cite{aytekin2018probabilistic}. In this paper, we argue that a graph-based learning approach can also be used to find image regions with similar color properties and to weight their contribution in estimating the global illumination.

In graph-based learning, an image is represented by a graph \textit{G = (V, E)}, where each node $v_i \in \textit{V}$ is represented by a vector $\textbf{x}_i$, e.g., color of a pixel  or a super-pixel in the image, and an edge $(v_i, v_j) \in E$ connects nodes $v_i$ and $v_j$. Each edge has a weight $s_{ij}$, which typically represents the similarity of nodes $v_i$ and $v_j$. Usually, graph-based approaches rely on two matrices, a similarity matrix $\textbf{S}= \{s_{ij}\}$ and a diagonal matrix $\textbf{Q}$, which can be used either for regularization or to encode prior knowledge and constraints depending on the problem. In a segmentation task, where the main goal is to divide an image into different regions possessing similar semantic properties, the problem becomes equivalent to dividing the graph based on paths with a minimal combined cost, where the cost for each edge is defined by the corresponding similarity $s_{ij}$. The similarity is defined based on the desired properties for the given task \cite{vicente2008graph,shi2000normalized}. For example, in a saliency estimation task, such as in \cite{aytekin2018probabilistic}, the main goal is to estimate the saliency of each pixel or region. In \cite{aytekin2018probabilistic}, the similarity matrix $\textbf{S}$ is constructed based on spatial and color-wise distances and $\textbf{Q}$ encodes prior knowledge that pixels at the borders of the image
are not likely to be salient.

% classification \cite{iosifidis2016multi,bai2012graph}, semi-supervised learning \cite{camps2007semi}, tracking \cite{gomila2003graph} and segmentation \cite{vicente2008graph,aytekin2017learning}. 

\section{Problem formulation } \label{Proposed}
We assume that the illumination color is in the convex hull of the pixel colors present in the scene. As a result, the color constancy problem becomes equivalent to estimating the probability $p(\textbf{x}_i)$ of each (super-)pixel $\textbf{x}_i$, i.e., each node $v_i$ of the graph \textit{G}, to contain information on the global illumination. The probability  $p(\textbf{x}_i)$ is higher when the region $\textbf{x}_i$ encodes more information about the illumination. Within this framework, we wish $\textbf{p}= \{ p(\textbf{x}_1),..., p(\textbf{x}_n) \} $ to satisfy properties associated with the illumination assumptions. We make the assumption that pixels with similar color encode similar information about the illumination and thus they are expected to have similar probabilities. In addition, we assume that dark pixels, i.e., pixels with low luminance, encode less information about the light illumination and thus they should have lower probabilities. Therefore, the estimation of $\textbf{p}$ can be formulated as an optimization problem with two terms. The first term enforces similar pixels to have similar probabilities and the second one encodes the prior knowledge and reflects the assumption that darker spots are usually less affected by the illumination. The joint optimization problem can be expressed as follows:
\begin{equation}
\label{aaa}
   \left\{
                \begin{array}{ll}
 \underset{\textbf{p}} {\mathrm{argmin}}\hspace{0.07cm}  \sum_{i} p(\textbf{x}_i)^2 q_i  +  \frac{1}{2} \sum_{i,j} ( p(\textbf{x}_i) -  p(\textbf{x}_j))^2 s_{ij} , \\

 \mathrm{subject}\hspace{0.2cm}  \mathrm{to} \hspace{0.5cm}   \sum_{i} p(\textbf{x}_i) = 1.\\
   \end{array}
              \right. 
\end{equation}

The first term lessens the probability of a region $\textbf{x}_i$ if there is prior information that this region does not encode information about the illumination, e.g., is dark.  If $q_i$ corresponding to $\textbf{x}_i$ is high  the system will tend to give a lower weight to $\textbf{x}_i$ to ensure the minimization of the total sum. The weight $s_{ij}$ encodes the similarity between two (super-)pixels $\textbf{x}_i$ and $\textbf{x}_j$ and thus if $s_{ij}$ is high, the resulting probabilities $p(\textbf{x}_i)$ and $p(\textbf{x}_j)$ will be similar.  $ \sum_{i} p(\textbf{x}_i) = 1$ is a normalization constraint ensuring that the result of the optimization problem (2) is a probability density function.

The above optimization problem can be framed as a graph-based learning problem with a similarity matrix formed by $\textbf{S}_{ij} = s_{ij}$  and $\textbf{Q}= diag(q_1, ...,q_n)$ and is similar to the problem defined in \cite{aytekin2018probabilistic}. Thus, the graph-based formulation of the problem (\ref{aaa})  can be expressed as follows:

\begin{equation}
 \left\{
                \begin{array}{ll}
 \underset{\textbf{p}}{\mathrm{argmin}}\hspace{0.07cm}  \textbf{p}^t \textbf{H} \textbf{p}, \\

 \mathrm{subject}\hspace{0.2cm}  \mathrm{to} \hspace{0.5cm}  \textbf{p}^t \textbf{1} = 1    ,\\
   \end{array}
              \right.  
\end{equation}
where $\textbf{H} = \textbf{D} - \textbf{S} + \textbf{Q}$ and $\textbf{D}$  the graph degree matrix defined as $\textbf{D}_{ii} = \sum_j \textbf{S}_{ij}$. The optimization problem in (3) has one global optimum solution which can be obtained as
\begin{equation}
  \textbf{p}^* = \textbf{H}^{-1}  \textbf{1}.  
\end{equation}
Having computed \textbf{p}*, the illumination estimation can be computed as a weighted sum of the pixels color in the input image:
\begin{equation}
   \textbf{I}^{est} = \sum_i \textbf{x}_i p^*(\textbf{x}_i).
\end{equation}

For the construction of $\textbf{Q}$ and $\textbf{S}$, we rely on the \textit{Lab} color space. The \textit{Lab} space is intuitive and appropriate in a color constancy context as it was designed to be perceptually uniform with respect to human color vision, meaning that the same amount of numerical change in these values corresponds to about the same amount of visually perceived change, and with respect to a given white point, it is device-independent. That is, it defines colors independent of how they are created or displayed. We propose two variants of $\textbf{Q}$ in Eq. (6-7). The first, defined in Eq. (6), is a fixed threshold, while the second one, defined in Eq. (7), has a linear form which gives larger values to pixels with higher lightening:

\begin{equation}
  \textbf{Q}_1(\textbf{x}_i) = \left\{
                \begin{array}{ll}
 0.1\hspace{0.3cm} if \hspace{0.1cm} L(\textbf{x}_i)< q_0, \\

 0 \hspace{0.3cm}else \\
  \end{array}
              \right.  
\end{equation}

\begin{equation}
  \textbf{Q}_2(\textbf{x}_i) =  0.1  \frac{ L_{max} - L(\textbf{x}_i) }{L_{max} - L_{min} }.
\end{equation}
$L(x)$ is the Lightness channel of \textit{lab} color space.  $L_{max}$ and $L_{min}$ are the maximum and the minimum values of  $L(x)$ in the input image, respectively, and  $q_0$ is a threshold  value for $\textbf{Q}_1$. The similarity function, used to calculate the elements of $\textbf{S}$, reflects the assumption that pixels with similar colors are similarly affected by the illumination. It can be defined as follow: 
\begin{equation}
  S(\textbf{x}_i,\textbf{x}_j)=  1 /(\epsilon +  || lab(\textbf{x}_i) -  lab(\textbf{x}_j)||_2/ \gamma ),
\end{equation}
where $lab(\textbf{x}_i)$ is the color of the pixel $\textbf{x}_i$ in the lab space, $\epsilon$ is a small constant, and $\gamma$ is a scaling parameter. It should be noted that different forms of $\textbf{Q}$ and $\textbf{S}$ can be used and more sophisticated color constancy prior knowledge can be integrated in the future.

\section{Experiments and Discussion }
We evaluated the proposed method on the recently released 
INTEL-TAU dataset \cite{laakom2019intel}. The dataset contains 7022 images in total captured using three different camera models, Canon 5DSR, Nikon D810, and Sony IMX135. The dataset is the largest available high-resolution dataset for illumination estimation research.
% \subsubsection*{NUS-8}  
% NUS-8 \cite{NUS} is a publicly available dataset, containing 1736 raw images from eight different camera models. Each camera has about  210  images.
Due to the high dimension of the images (1080p), we abstract the image as (super)-pixels \cite{achanta2010slic}. Each image is represented by  800 super-pixels. The parameter $\gamma $ defined in Eq. (8) is set to $2$. Another hyperparameter is the percentage of pixels used in the final estimate Eq. (5). We rely on 1\% of the pixels with the highest probability values $\textbf{p}^*(\textbf{x}_i)$ computed using Eq. (4) to obtain the final estimate of the illumination. The threshold $q_0$ in Eq. (6) is set to $10\%$ of $L_{max}$.

We report the mean of the top 25\%, the mean, the median, Tukey's trimean, and the mean of the worst 25\% of the Recovery angular error  $e_{recovery}$ \cite{21}  and the `Reproduction angular error' $e_{reproduction}$ \cite{finlayson2014reproduction} between the ground truth illuminant and the estimated illuminant: 
\begin{equation}
       \text{$e_{recovery}$}(\textbf{I}^{gt},\textbf{I}^{est})= \cos^{-1} ({ \frac{ \textbf{I}^{gt} \textbf{I}^{est}}{\| \textbf{I}^{gt} \| \|\textbf{I}^{est} \| } }) 
\end{equation}

\begin{equation}
     \text{$e_{reproduction}$}(\textbf{I}^{gt},\textbf{I}^{est})= \cos^{-1} ({ \frac{ \textbf{I}^{gt}/ \textbf{I}^{est} \hspace{2mm} \textbf{w}}{\| \textbf{I}^{gt} /\textbf{I}^{est} \|  \sqrt3} }), 
\end{equation}
where $\textbf{I}^{gt}$ is the ground truth illumination for an image, $\textbf{I}^{est}$ is the estimated illumination, $/$ is the element-wise division operator, and $evaluated$ is defined as the unit vector, i.e., $\textbf{w} = [1 , 1, 1 ]^t$. 

\begin{table*}[h]
	\caption{ Results of benchmark methods on INTEL-TAU Dataset.}
		\label{tab:intelt}

\begin{tabular}{l|lllll||lllll}
                    & \multicolumn{5}{c||}{Recovery angular error}                                                       & \multicolumn{5}{c}{Reproduction angular error}                                                  \\ \hline
Method              & Best \newline 25\% & Mean & Med. & Tri. & W. \newline 25\% & Best \newline 25\% & Mean & Med. & Tri. & W. \newline 25\% \\
\hline
Grey-World \cite{4}         & 1.0                               & 4.9  & 3.9  & 4.1  & 10.5                            & 1.2                               & 6.1  & 4.9  & 5.2  & 13.0                            \\
White-Patch   \cite{2}      & 1.4                               & 9.4  & 9.1  & 9.2  & 17.6                            & 1.8                               & 10.0  & 9.5  & 9.8  & 19.2                            \\
Grey-Edge    \cite{5}       & 1.0                               & 5.9  & 4.0  & 4.6  & 13.8                            & 1.2                               & 6.8  & 4.9  & 5.5  & 13.5                           \\
2nd order Grey-Edge \cite{5}  & 1.0                               & 6.0  & 3.9  & 4.8  & 14.0                            & 1.2                               & 6.9  & 4.9  & 5.6  & 15.7                            \\
Shades-of-Grey   \cite{shades}   & 0.9                               & 5.2  & 3.8  & 4.3  & 11.9                             & 1.1                               & 6.3  & 4.7  & 5.1  & 13.9                            \\
Cheng et al. 2014  \cite{NUS}   & 0.7 & 4.5  & 3.2  & 3.5  & 10.6                            & \cellcolor[gray]{0.8}0.9                               & 5.5  & 4.0  & 4.4  & 12.7                            \\
Weighted Grey-Edge  \cite{5} & \cellcolor[gray]{0.8}0.8                               & 6.1  & 3.7  & 4.6  & 15.1                            & 1.1                               & 6.9  & 4.5  & 5.4  & 16.5                            \\
Yang et al. 2015  \cite{yang2015efficient}  & \cellcolor[gray]{0.65}0.6                               & \cellcolor[gray]{0.5}3.2  & \cellcolor[gray]{0.5}2.2  & \cellcolor[gray]{0.5}2.4  & \cellcolor[gray]{0.5}7.6                             & \cellcolor[gray]{0.65}0.7                               & \cellcolor[gray]{0.5}4.1  & \cellcolor[gray]{0.5}2.7  & \cellcolor[gray]{0.5}3.1  & \cellcolor[gray]{0.5}9.6                            \\
Greyness Index 2019 \cite{qian2019finding} & \cellcolor[gray]{0.5}0.5                               & \cellcolor[gray]{0.65}3.9    & \cellcolor[gray]{0.65}2.3  & \cellcolor[gray]{0.65}2.7  & \cellcolor[gray]{0.8}9.8                             & \cellcolor[gray]{0.5}0.6                               & \cellcolor[gray]{0.65}4.9  & \cellcolor[gray]{0.65}3.0  & \cellcolor[gray]{0.65}3.5  & \cellcolor[gray]{0.8}12.1    

             \\
Color Tiger \cite{banic2017unsupervised}
& 1.0                               & 4.2    & 2.6  & 3.2  & 9.9                             & 1.1                              & 5.3  & \cellcolor[gray]{0.8}3.3  & 4.1  & 12.7   \\
\hline

PCC\_Q1 & \cellcolor[gray]{0.65}0.6                               & \cellcolor[gray]{0.8}4.1    & 2.6  & 3.0  & 10.1                             & \cellcolor[gray]{0.65}0.7                              & 5.3  & 3.7  & 4.2  & 12.3 \\

PCC\_Q2 & \cellcolor[gray]{0.65}0.6                               & \cellcolor[gray]{0.65}3.9    & \cellcolor[gray]{0.8}2.4  & \cellcolor[gray]{0.8}2.8  & \cellcolor[gray]{0.65}9.6                             & \cellcolor[gray]{0.65}0.7 & \cellcolor[gray]{0.8}5.1& 3.5  & \cellcolor[gray]{0.8}4.0  & \cellcolor[gray]{0.65}11.9 

\end{tabular}
\end{table*}

%V2current_gamma =2; %gamma(ii) ;
%supp = 800  ;% pixels(jj) ;
%v= 99;

We compare the proposed method with the state-of-the-art unsupervised methods Grey-World \cite{4}, White-Patch   \cite{2}, Grey-Edge variants  \cite{5}, Shades-of-Grey  \cite{shades}, Weighted Grey-Edge \cite{5}, Greyness Index \cite{qian2019finding}, Color Tiger \cite{banic2017unsupervised}, and methods proposed in \cite{NUS} and \cite{yang2015efficient}. The performances of these methods are reported in Table \ref{tab:intelt}.  We also provide results for both variants of our approach, namely PCC\_Q1 and PCC\_Q2, obtained using the prior knowledge matrices $\textbf{Q}_1$ and $\textbf{Q}_2$, respectively. 

In terms of recovery angular error, our method achieves better overall performance than Grey-World, White-Patch, Grey-Edge, second order Grey-Edge, Weighted Grey-Edge, method in \cite{NUS}, and Color Tiger. The method in \cite{yang2015efficient} achieves the best overall results in terms of Recovery angular error statistics.   In terms of Reproduction angular error, our method outperforms Color Tiger in the mean of Best 25\% and worst 25\% while Color Tiger yields a better median performance. Both Greyness Index and method in \cite{yang2015efficient} outperform our approach mainly in terms of median and mean errors.  We note that the linear second variant of $\textbf{Q}$ outperforms the  static first variant.

\begin{figure}[t]
\centering
\includegraphics[width=8.8cm]{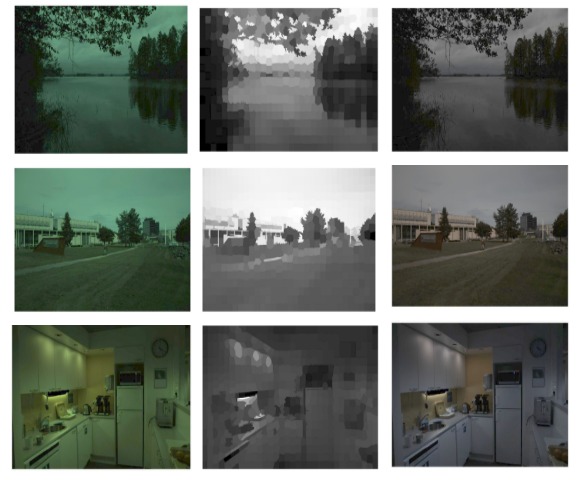}
% where an .eps filename suffix will be assumed under latex, 
% and a .pdf suffix will be assumed for pdf latex; or what has been declared
% via \DeclareGraphicsExtensions.
\caption{Visual results on INTEL-TAU using our approach. From left to right, our input images from INTEL-TAU, the probability distribution mask obtained by PCC, our corrected image.
}
\label{fig_1}
\end{figure}

Figure \ref{fig_1} illustrates some visual results using the proposed approach. The middle column shows the weight masks obtained using the proposed approach. We see that depending on the scene characteristics, our approach generates different weight masks. It should be noted that using our approach one can generate local estimates per pixel by simply multiplying each pixel by its corresponding weight.  

\section{Conclusion}
In this paper, a novel unsupervised color constancy approach is proposed. In this approach, we defined a probability-based method for estimating illumination of a scene using a graph-based representation of the image. We showed that the illumination estimation problem can be solved by estimating a probability density function defined over the image. To this end, we proposed two intuitive assumptions that enable us to define the graph connectivities. The proposed algorithm was tested on  INTEL-TAU dataset, where we achieved comparable results to state-of-the-art methods. Future work includes proposing different variants of the unsupervised graph-based approach and incorporating other color constancy assumptions into the framework.

\section*{ACKNOWLEDGEMENT}

This  work  was  supported  by  the  NSF-Business Finland  Center  for
Visual and Decision Informatics (CVDI) project AMALIA 2019.
\bibliographystyle{IEEEbib}
\bibliography{strings,refs}

\end{document}